\documentclass[letterpaper]{article} 
\usepackage[]{aaai25}  
\usepackage{times}  
\usepackage{helvet}  
\usepackage{courier}  
\usepackage[hyphens]{url}  
\usepackage{graphicx} 
\usepackage{natbib}  
\usepackage{caption} 
\usepackage{algorithm}
\usepackage{algorithmic}
\usepackage{amsmath}
\usepackage{mathrsfs}
\usepackage[table]{xcolor}
\usepackage{amssymb}
\usepackage{booktabs} 
\usepackage{multirow} 
\usepackage{colortbl} 
\usepackage{rotating}
\usepackage{newfloat}
\usepackage{listings}
\DeclareCaptionStyle{ruled}{labelfont=normalfont,labelsep=colon,strut=off} 
\floatstyle{ruled}
\newfloat{listing}{tb}{lst}{}
\floatname{listing}{Listing}
\title{Vision-Language Navigation with Continual Learning}

\author{
    Zhiyuan Li\textsuperscript{\rm 1, \rm 2},
    Yanfeng Lv\textsuperscript{\rm 1, \rm 2},
    Ziqin Tu\textsuperscript{\rm 1, \rm 2},
    Di Shang\textsuperscript{\rm 1, \rm 2},
    Hong Qiao\textsuperscript{\rm 1, \rm 2}
}
\affiliations{
    \textsuperscript{\rm 1}State Key Laboratory of Multimodal Artificial Intelligence Systems,\\
    Institute of Automation, Chinese Academy of Science (CASIA) Beijing 100190, China\\
    \textsuperscript{\rm 2}University of Chinese Academy of Sciences (UCAS), Beijing 100049, China \\
    lizhiyuan2022@ia.ac.cn, yanfeng.lv@ia.ac.cn

}

\usepackage{bibentry}

\begin{document}

\maketitle

\begin{abstract}
Vision-language navigation (VLN) is a critical domain within embedded intelligence, requiring agents to navigate 3D environments based on natural language instructions. Traditional VLN research has focused on improving environmental understanding and decision accuracy. However, these approaches often exhibit a significant performance gap when agents are deployed in novel environments, mainly due to the limited diversity of training data. Expanding datasets to cover a broader range of environments is impractical and costly.
%
We propose the Vision-Language Navigation with Continual Learning (VLNCL) paradigm to address this challenge. In this paradigm, agents incrementally learn new environments while retaining previously acquired knowledge. VLNCL enables agents to maintain an environmental memory and extract relevant knowledge, allowing rapid adaptation to new environments while preserving existing information.
We introduce a novel dual-loop scenario replay method (Dual-SR) inspired by brain memory replay mechanisms integrated with VLN agents. This method facilitates consolidating past experiences and enhances generalization across new tasks. By utilizing a multi-scenario memory buffer, the agent efficiently organizes and replays task memories, thereby bolstering its ability to adapt quickly to new environments and mitigating catastrophic forgetting.
%
Our work pioneers continual learning in VLN agents, introducing a novel experimental setup and evaluation metrics. We demonstrate the effectiveness of our approach through extensive evaluations and establish a benchmark for the VLNCL paradigm. Comparative experiments with existing continual learning and VLN methods show significant improvements, achieving state-of-the-art performance in continual learning ability and highlighting the potential of our approach in enabling rapid adaptation while preserving prior knowledge.



\end{abstract}

%

\section{Introduction}

Vision-Language Navigation (VLN) \cite{anderson2018vision} is crucial for the embedding intelligence field. The agent follows natural language instructions and moves around in the 3D environment. By integrating natural language processing, visual perception, and decision-making, the agent could navigate to the destination. Most research in VLN focuses on environment understanding ability improvement \cite{hong2020language} and accuracy of target decision policy \cite{hao2020towards}. 
While these advancements have significantly improved VLN performance, there is still a critical issue: the generalization of agents to diverse unseen scenes, which is essential for real-world applications.
In practical scenarios, agents must continually adapt to new environments while retaining the knowledge acquired from previous tasks.
The significant performance gap between seen and unseen \cite{anderson2018vision} environments underscores this challenge. The primary cause of this issue is the limited availability of diverse environmental data, which constrains the agents' ability to generalize effectively \cite{zhang2020diagnosing}.
Yet, massively expanding the dataset with various environments is unrealistic and expensive \cite{shah2023lm}. Therefore, we consider an alternative approach by introducing the continual learning (CL) framework. 
This framework enables the agents to incrementally learn and adapt to new environments while retaining the knowledge acquired from previous tasks \cite{srinivasan2022climb}. By using this strategy, we aim to enhance the generalization capabilities of VLN agents, making them more robust and effective in real-world applications where they must navigate an ever-changing array of environments.

To enable the VLN agent to accumulate knowledge from tasks, handling the challenge known as catastrophic forgetting \cite{french1999catastrophic} is vital. We combine it with vision-language navigation tasks to introduce the Vision-Language Navigation with Continual Learning (VLNCL) paradigm. The agent must continuously accumulate information and maintain former knowledge by motivating the agent with new tasks. That means a balance between stability and plasticity \cite{kim2023achieving}. Furthermore, considering that real-world tasks often occur within the same environment simultaneously, we split the tasks by scene to raise them to the agent. This way, tasks are divided into different domains.


\begin{figure*}
\centering
\includegraphics[width=0.9\textwidth]{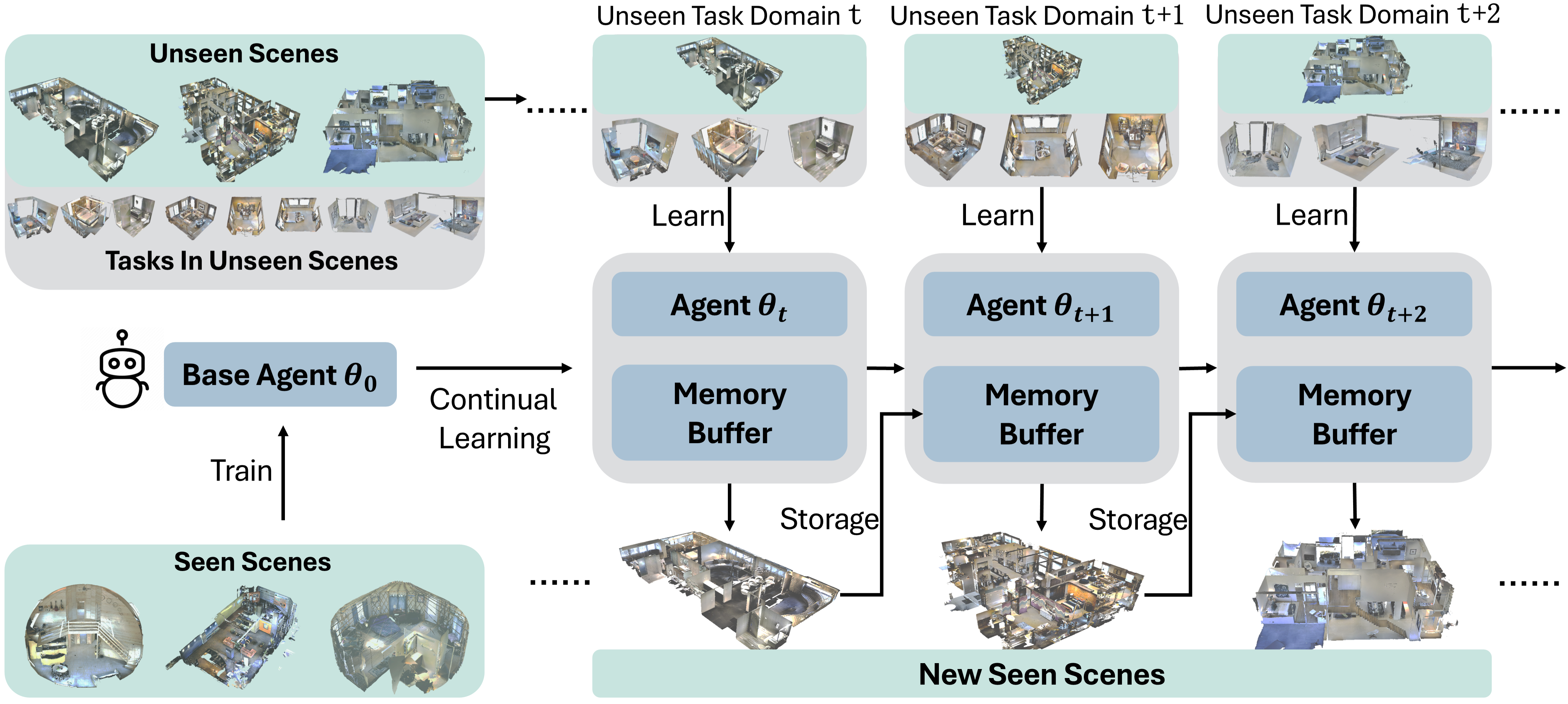}
\caption{The pipeline of Vision-Language Navigation with Continual Learning (VLNCL). The agent is trained in the seen dataset to achieve the base agent. When encountering various unseen tasks, the VLNCL paradigm requires the agent to continuously learn from new tasks while not forgetting former scene information.}
\end{figure*}

Building on this foundation, we propose the dual-loop scenario replay vision-language navigation agent (Dual-SR) as a novel method for VLNCL. Inspired by the mechanism of memory replay in the resting brain \cite{zhong2024waking}, we designed a dual-loop memory replay framework to enable the model to consolidate earlier scenario memories while balancing new task learning. Randomly replaying scenario memory from the memory buffer of the agent brings a former task memory bias while the inner loop weight updates learning new tasks \cite{rolnick2019experience}. 
Then, the agent applies the meta-learning-based outer loop weight updates to balance the new and old weights while alleviating the overfitting problem caused by the single scenario task data \cite{javed2019meta}.
Additionally, to effectively retain diverse environmental knowledge, we design a memory buffer based on task domains that allow agents to store and replay memories from different scenes.
To assess VLNCL agents, we propose two metrics: Unseen Transfer (UT) for evaluating knowledge transfer and Seen Transfer (ST) for assessing continual learning. UT measures generalization by testing the agent in new scenes with unseen tasks. ST evaluates continual learning by applying the agent to all seen task domains after training. 


Making the VLN agent able to perform continual learning in new tasks brings the advantage of increasing task performance and increasing task generalization. Comparative experiments were also conducted extensively with several CL methods used in other fields and previous VLN agents to verify the advances of our approach. Experiments show state-of-the-art performance in continual learning ability.

To summarize the contribution of this work:

\begin{itemize}
    \item We introduce the Vision-Language Navigation with Continual Learning (VLNCL) paradigm and metrics, enabling VLN agents to adapt to new unseen environments while retaining prior knowledge for improving generalization. 
    \item We propose Dual-loop Scenario Replay (Dual-SR), a novel approach inspired by the brain memory system, which enables continual learning in VLN agents. Experimental results showed a 16\% success rate rise compared with the base model. 
    \item We design a multi-scenario memory buffer that organizes task memories by environment type, facilitating rapid adaptation and balancing tasks across scenarios.
    \item Our work focuses on continual learning in VLN agents, reduces catastrophic forgetting, and improves knowledge transfer, setting a benchmark in the field. Comparative experiments also confirm its robustness and effectiveness. 
\end{itemize}

\section{Related Work}

\subsection{Vision-and-Language Navigation}

Vision-and-Language Navigation (VLN) aims to develop intelligent agents capable of interacting with humans using natural language, perceiving the environment, and executing real-world tasks. This field has attracted significant attention across natural language processing, computer vision, robotics, and machine learning. Anderson et al. \cite{anderson2018vision} laid the foundation by introducing the Room-to-Room (R2R) dataset, where agents navigate virtual indoor environments using a simulator. Building on R2R, researchers quickly established other VLN benchmarks like R4R \cite{jain2019stay} and RxR \cite{ku2020room}.

For outdoor navigation, Touchdown \cite{chen2019touchdown} is a crucial benchmark where agents navigate a simulated New York City street view. Conversational navigation tasks, such as CVDN \cite{thomason2020vision} and HANNA \cite{nguyen2019help}, enable agents to interact with humans to aid navigation. Remote object navigation tasks, including REVERIE \cite{qi2020reverie} and SOON \cite{zhu2021soon}, require agents to infer object locations and identify them based on language instructions.

To address cross-modal alignment, PTA \cite{landi2021multimodal} leverages CNN-based visual features, while HAMT \cite{chen2021history} incorporates long-term history into multimodal decision-making using a hierarchical vision transformer. Kerm \cite{li2023kerm} models relationships between scenes, objects, and directions using graphs. For navigation decisions, Dreamwalker \cite{wang2023dreamwalker} employs reinforcement learning with an intrinsic reward for cross-modal matching. Transformer-based models \cite{zhao2022target} have also gained popularity, integrating visual and linguistic information to enhance decision-making.

Despite these advances, introducing new environments requires retraining on seen environments to prevent catastrophic forgetting, leading to high costs and limited adaptability. Thus, enabling VLN agents with continual learning capabilities is crucial.

\subsection{Continual Learning}


Continual learning is vital for agents to acquire new tasks without forgetting prior ones. Compared with traditional machine learning, which relies on static datasets, continual learning processes sequential data streams, making earlier data inaccessible. First introduced by Thrun \cite{thrun1998lifelong}, recent research has focused on mitigating catastrophic forgetting \cite{mccloskey1989catastrophic}. Approaches in this field generally fall into three categories: rehearsal-based, regularization-based, and parameter isolation methods.

Rehearsal-based methods, like experience replay, retain data from previous tasks to update the model during new task training \cite{chaudhry2019tiny}. Techniques by Lopez-Paz \cite{lopez2017gradient} and Chaudhry \cite{chaudhry2018efficient} use data replay to prevent gradient conflicts. While combined with meta-learning and experience replay, these methods help optimize network features and reduce conflicts between tasks \cite{le2019learning, ho2023prototype}. Capacity expansion approaches \cite{liu2022few, ramesh2021model} and online meta-learning \cite{gupta2020look} offer additional strategies to mitigate forgetting.

In the VLN domain, continual learning research is still developing. Jeong et al. \cite{jeong2024continual} recently proposed rehearsal-based methods like "PerpR" and "ESR" to alleviate the catastrophic forgetting problem that occurs when new tasks are continuously input. This work raises the possibility of combining continual learning with visual-language navigation. However, this work only focuses on alleviating catastrophic forgetting but ignores the necessity of continuously improving the generalization ability of intelligent agents in unfamiliar environments in real application scenarios. Further refinement of the continual learning framework and evaluation metrics is needed.

\section{Method}

\subsection{Setting of Vision-Language Navigation}

Vision-language navigation (VLN) involves an agent navigating photorealistic indoor environments based on visual inputs $\mathbf{V}$ and language instructions $\mathbf{I}$. This problem can be modeled as a Markov Decision Process, where the agent's state at each time step $s_t$ represents its visual observation and position in the environment. The agent’s policy $\pi$ maps these states to actions $a_t$ that lead the agent toward the desired goal. The learning process involves minimizing the expected loss over a trajectory $\tau$ generated by the agent's policy $\pi$:
\begin{equation}
    \mathbb{E}_{\tau \sim \pi} \left[ \sum_{t=0}^{T} L(s_t, a_t) \right]
\end{equation}
where $L(s_t, a_t)$ is the loss at each time step. The integration of vision and language occurs by projecting $\mathbf{I}$ and $\mathbf{V}$ into a common feature space, creating a joint embedding space.
This embedding allows the agent to align visual cues with linguistic references, enabling accurate navigation.


\subsection{Formulation of Vision-Language Navigation with Continual Learning}

In practical applications, agents must adapt to unseen environments while retaining knowledge from previously encountered scenes. To this end, we adopt a continual learning approach for vision-language navigation (VLN), where the validation dataset is split into multiple data streams, simulating different task domains. Each data stream consists of tasks from a specific scene, allowing the agent to learn across various environments without forgetting previously learned tasks sequentially.

We partition the dataset into $d$ task domains, denoted as $\mathcal{TD} = \{td_1, td_2, \dots, td_d\}$, where each $td_i$ represents a distinct scene and task domains are considered independent. Each task domain $td_i$ is modeled as a distribution $D_{td_i}$ in this setting.  We reformulate the loss function from the VLN setting to the VLNCL setting as follows:
\begin{equation}
\mathbb{E}_{\tau \sim \pi} \left[ \sum_{i=1}^{d} \sum_{t=0}^{T_i} \mathcal{L}(s_t^{(i)}, a_t^{(i)}) \right],
\end{equation}
where $T_i$ is the time horizon for task domain $td_i$, $s_t^{(i)}$ and $a_t^{(i)}$ are the agent's state and action in domain $td_i$ at time $t$, respectively. The continual learning framework generalizes this to multiple task domains, allowing the agent to adapt progressively to new environments. This framework improves performance in unseen environments by balancing the trade-off between retaining knowledge and promoting generalization.

\subsection{Dual-loop Scenario Replay}

In the VLNCL setup, agents must minimize forgetting and improve transfer learning by leveraging prior knowledge to improve performance on current and former tasks. However, most existing VLN agents struggle with continual learning, unlike humans, who efficiently learn from a few examples by integrating sensory input with long-term memory \cite{goelet1986long}. The human brain continuously extracts and stores knowledge, reinforcing long-term memory through replay during rest \cite{dewar2012brief}. Inspired by this, we propose the Dual-loop Scenario Replay Continual Learning (Dual-SR) algorithm for VLN agents. This algorithm simulates working memory as an inner loop and long-term memory as an outer loop, creating two weight update loops to balance prior and current task information while enhancing generalization.

\begin{figure}
\includegraphics[width=0.48\textwidth]{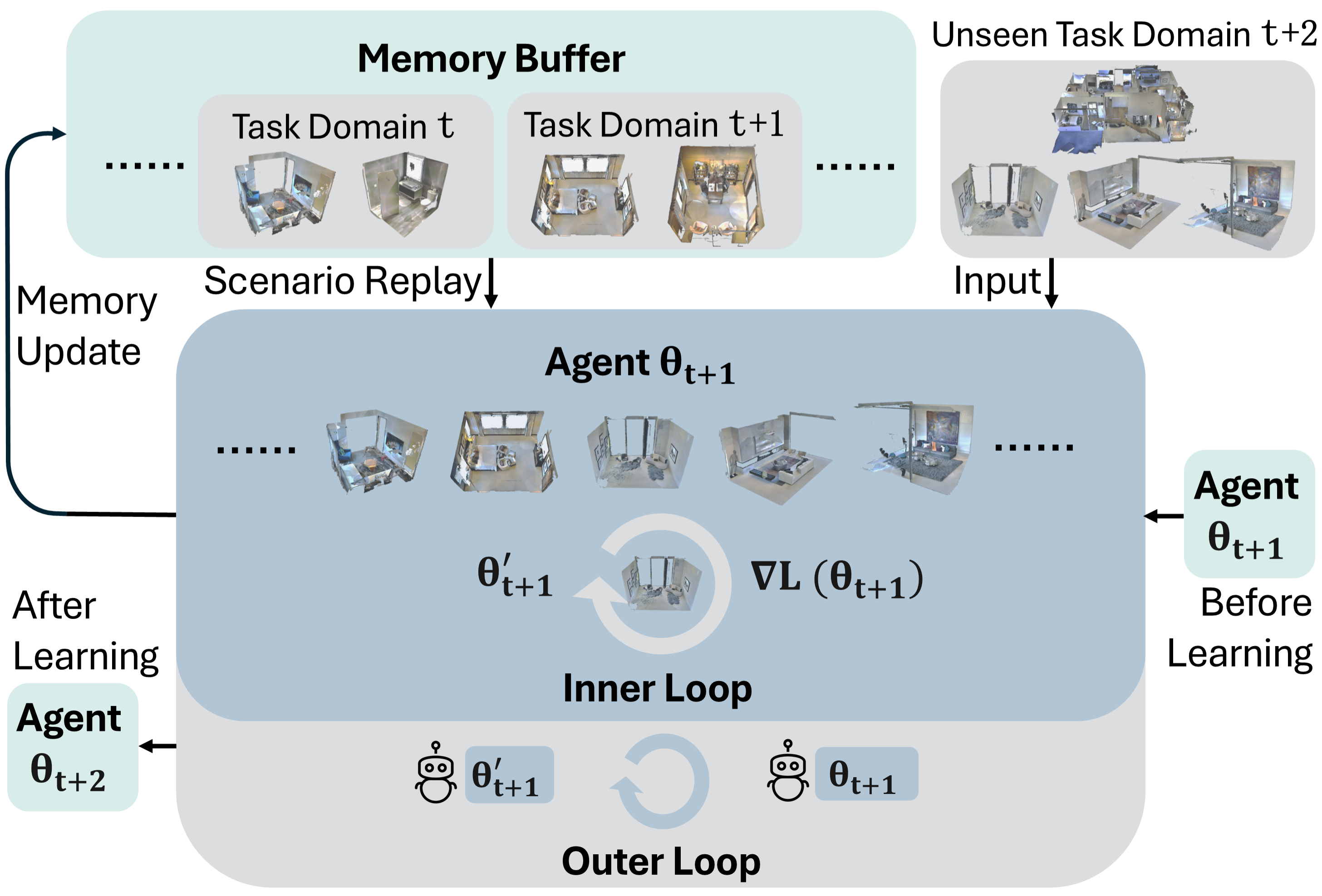}
\caption{The overview of Dual-loop Scenario Replay (Dual-SR) algorithm for the VLN agent. 
When the agent receives new, unseen task domain data, it randomly replays former tasks from the memory buffer to update the inner loop. After the new task domain learning process finishes, the agent performs the outer loop update to balance agent parameters.}
\end{figure}

In the VLNCL setup, agents might encounter an overfitting problem due to the limited samples. Hence, we leverage the meta-update mechanism in the Reptile algorithm \cite{nichol2018first} to mimic long-term memory formation. The reptile algorithm is equivalent in effectiveness to MAML \cite{genzel2015role}, which provides a means for models to acquire a standard structure from the current task domain, enabling them to adapt to other similar new tasks quickly. Thus, we can maximize generalization ability rather than data fitting. By that, the outer loop can improve the generalization capacity of agents. In the outer loop, the update of weight can be defined as:
\begin{equation}
\label{eq:metaupdate}
  \theta= \theta + \beta*(\theta^{'}-\theta)
\end{equation}
where $\theta$, $\theta^{'}$, and $\beta$ separately denote weights of the model before the inner cycle, weights after it, and meta-learning rate. 


To mimic the brain's abstraction and consolidation of long-term memories \cite{goelet1986long}, we designed the inner loop by simulating the memory retrieval in working memory. The agent can train with stable data selected randomly from previous task domains by maintaining the buffer and replaying old samples in the inner loop. Applying the memory buffer can ensure each former task is equally likely to be selected in the buffer. The newly received samples are combined with randomly selected old samples from the buffer to form a mini-batch, which is then used for meta-learning. 
In the inner loop, the update of the model can be defined as: 
\begin{equation}
\label{eq:inner-loop}
    U_{k}(\theta)=\theta-\alpha\circ\nabla L_{(a_t \mid \mathbf{V}, \mathbf{I})}(\theta)
\end{equation}
where $U_{k}(\theta)$ is a update by learning $(a_t \mid \mathbf{V}, \mathbf{I})$ and $\alpha$ are parameters of the meta-learner to be learned, and $\circ$ denotes element-wise product. Specifically, $\alpha$ is a vector of the same size as $\theta$ that decides both the update direction and learning rates. 

Meta-updates can extract common structures learned across tasks, thereby enhancing the knowledge transfer capabilities of agents. The VLNCL setup exposes agents to a dynamic and unpredictable data stream. This procedure necessitates that agents adapt and perform effectively across an evolving array of tasks without following a predefined sequence. This approach differs significantly from traditional methods that require partitioning a fixed dataset into multiple batches for a set number of tasks. To address this, we implement experience replay within the inner loop. By storing task indices, the agent can revisit and leverage previously learned tasks when faced with new ones. 

This approach contrasts with traditional replay-based methods, which indiscriminately use memory across all tasks. Our method randomly replays scenarios within each task domain to ensure balance. Additionally, we introduce the memory buffer size $Z$. 
When a task belongs to a previous task domain and the task ID $t$ is a multiple of $Z$, the agent updates the memory buffer $M$ by replacing one of the tasks in the corresponding domain with the current task.
The agent efficiently manages memory size by updating scenario memory between task domains, even when handling many tasks. This strategy also encourages the model to prioritize tasks inspired by working memory principles.

\begin{algorithm}[ht]
\caption{Dual-loop Scenario Replay (Dual-SR)}
\label{alg:dual_sr}
\textbf{Input}: Base parameters $\theta$, buffer $M$, learning rates $\alpha$, $\beta$, replay number $rs$. \\ 
\textbf{Output}: Updated parameters $\theta$. \vspace{-1.2em} \\
\begin{algorithmic}[1]
    \STATE Initialize $\theta \leftarrow \theta_0$, $M \leftarrow \emptyset$
    \FOR{each task domain $s$}
        \FOR{each task batch $B_t \in s$}
            \STATE Random sampling $B_M = sample(rs, M)$
            \STATE Sample batches $B_t$ from $s$ and $B_M$ from $M$
            \STATE Update $\theta \leftarrow \theta - \alpha \circ \nabla L(\theta; B_t \cup B_M)$
        \ENDFOR
        \STATE Update $\theta \leftarrow \theta + \beta * (\theta' - \theta)$
        \STATE Update $M$ with $s$
    \ENDFOR
    \STATE \textbf{return} $\theta$
\end{algorithmic}
\end{algorithm}



\subsection{Structured Transformer VLN Agents with Continual Learning}

Building upon the Dual-SR algorithm, we employ a cross-modal structured transformer \cite{zhao2022target, chen2021history, lu2019vilbert} as the planner to enhance the performance of the VLN agent in continual learning settings. The Dual-SR algorithm provides the foundation for this approach by balancing integrating new information and retaining prior knowledge. At each navigation step $t$, the model processes five forms of tokens, global token $\mathbf{g}_{t-1}$, candidate target tokens $\mathbf{C} = \{\mathbf{c}_{t-1}^1, \mathbf{c}_{t-1}^2, \ldots, \mathbf{c}_{t-1}^q\}$, history tokens $\mathbf{H} = \{\mathbf{h}_{t-1}^1, \mathbf{h}_{t-1}^2, \ldots, \mathbf{h}_{t-1}^{t-1}\}$, encoded instruction tokens $\mathbf{I} = \{\mathbf{i}_0, \mathbf{i}_1, \ldots, \mathbf{i}_m\}$, and encoded vision tokens $\mathbf{V} = \{\mathbf{v}_t^1, \mathbf{v}_t^2, \ldots, \mathbf{v}_t^{n}\}$. The instruction tokens remain constant through time to reduce computation, and other tokens are updated based on previous time steps. The system initializes the global token as the sentence embedding $\mathbf{g}_{0} = i_0$.


To encode candidate target tokens, we apply the grid coding form to address the modeling of possible long-term target challenges in unseen scenes. Each cell center can represent a potential navigation target token by discretizing the environment into a $d \times d$ grid to cover the navigation area. Initially, candidate target tokens $\mathbf{c}_0^1, \mathbf{c}_0^2, \ldots, \mathbf{c}_0^q$ are created using the positional embedding of the targets, formulated as:
\begin{equation}
\mathbf{c}_0^i = f_{\phi_P}(\mathbf{s}_j) \cdot \mathbf{i}_0, \quad j \in \{1, 2, \ldots, q\}
\end{equation}
where $f_P$ is the positional encoder, $\mathbf{s}_i$ is the spatial location expressed by position coordinates, $\phi_P$ is the parameter of encoder, and $\mathbf{x}_0$ is the sentence embedding.


During navigation, these candidate target tokens are refined with new visual clues and instruction tokens to predict a more precise long-term target. The probability of each target being the navigation destination is calculated using a multi-layer perceptron (MLP) based target predictor:
\begin{equation}
\resizebox{0.9\hsize}{!}{$
    \begin{aligned}
		P(\mathbf{c}_t^i \mid \theta) = \text{softmax}\{\text{MLP}(\mathbf{c}_t^i \cdot \mathbf{g}_t)\}, \quad i \in \{1, 2, \ldots, q\}
	\end{aligned}
  $}
\end{equation}
where $\mathbf{g}_t$ is the global token.


The agent constructs and maintains a structured representation of the explored area with the transformer architecture to capture the structured environment layouts. At time step $t$, the model constructs a graph $\mathcal{S}_t$, where the nodes represent previously visited locations and the edges represent the navigability of those locations. We construct the history token $\mathbf{h}_t^t$ using panoramic view embedding, action embedding, temporal embedding, and positional embedding as follows:
\begin{equation}
	\mathbf{h}_t^t = f_{V}(\mathbf{v}_t^1, \ldots, \mathbf{v}_t^{n}) + f_{A}(\mathbf{r}_t) + f_{T}(t) + f_{P}(\mathbf{s}_t)
\end{equation}
where $f_V$ is a panoramic visual feature extractor, $\mathbf{r}_t = (\sin{\theta}, \cos{\theta}, \sin{\varphi}, \cos{\varphi})$ is the moving direction, $f_A$ is the action encoder, $f_T$ is the temporal encoder, and $f_P$ is the positional encoder.


The adjacency matrix $\mathbf{E}$ of history tokens at time step $t$ is defined such that if a navigation viewpoint $n_j$ is navigable from $n_i$, then $\mathbf{E}_{ij} = 1$; otherwise, $\mathbf{E}_{ij} = 0$. The attention mask matrix $\mathbf{M}$ controls the information flow among tokens, with a sub-matrix $\mathbf{M}_H$ for history tokens:
\begin{equation}
\mathbf{M}_H \leftarrow \mathbf{M}_H \ast \mathbf{E}
\end{equation}
where $\ast$ denotes element-wise multiplication.



The structured transformer enables the agent to access structured information of the past, allowing decisions from adjacent and previously visited locations. The local action space at time step $t$ is:
\begin{equation}
\mathcal{A}_t^L = \{\tau(\hat{\mathbf{v}}_t^1), \tau(\hat{\mathbf{v}}_t^2), \ldots, \tau(\hat{\mathbf{v}}_t^{k_t})\}
\end{equation}
and the global action space is:
\begin{equation}
\mathcal{A}_t^G = \{\tau(\hat{\mathbf{v}}_t^1), \ldots, \tau(\hat{\mathbf{v}}_t^{k_t}), \tau(\mathbf{h}_t^1), \ldots, \tau(\mathbf{h}_t^{t-1})\}
\end{equation}
where $\tau$ maps the token to its corresponding location. The probability of each possible action is:
\begin{equation}
\resizebox{0.87\hsize}{!}{$
    \begin{aligned}
		\pi(a_t \mid \theta) = \text{softmax}\{\text{MLP}(\tau^{-1}(a_t) \cdot \mathbf{g}_t)\}, \quad a_t \in \mathcal{A}_t^G
	\end{aligned}
  $}
\end{equation}

The optimization of the model involves both imitation learning (IL) loss $L_{IL}$ and reinforcement learning (RL) loss $L_{RL}$, alternating between teacher forcing (using ground truth actions) and student forcing (using actions sampled from the policy). To further consider the action chosen and target chosen, the history teacher loss $L_{HT}$ and target prediction loss $L_T$ are incorporated.
The history teacher loss is defined as:
\begin{equation}
L_{HT} = -\sum_{t=1}^T \log \pi(a_t \mid \theta)
\end{equation}
and the target prediction loss is:
\begin{equation}
L_T = -\sum_{t=1}^T \log P(\mathbf{c}_t^i \mid \theta)
\end{equation}
where the $i$-th target token is closest to the navigation destination.
The total loss function is given by:
\begin{equation}
L = \alpha_1 L_{IL} + \alpha_2 L_{RL} + \alpha_3 L_{HT} + \alpha_4 L_T
\end{equation}
where $\alpha_i$ are the loss coefficients.

After training the foundational model, the agent prompts continual learning inference within validation environments. The agent processes the task-domain-based data stream sequentially for the Val-Seen and Val-Unseen splits. The agent executes the inner loop based on the loss function specified in Equation \ref{eq:inner-loop} to iteratively update parameters $\theta$, achieved through continuous memory buffer updates and scenario replays. Upon completing the learning for the current task domain, the agent proceeds to perform the outer loop as described in Equation \ref{eq:metaupdate}.


The continual learning methodology equips the VLN agent to learn and adapt within complex environments, maintaining and enhancing knowledge across multiple tasks. The Dual-SR algorithm in structured transformers allows for effective navigation and adaptation capabilities in continually changing scenarios.



\section{Experiments}




\subsection{Experiment Setup}
The experiment adopts the VLNCL framework and divides the R2R dataset \cite{anderson2018vision} into distinct task domains to evaluate resistance to forgetting and knowledge transfer capabilities.
By sequentially inputting each task domain into the agent, we separately assess the average Seen Transfer (ST) and Unseen Transfer (UT) across each dataset split.

\begin{table}[ht]
  \caption{Number of task domains in each R2R dataset split.}
  \centering
  \tabcolsep=0.21cm
  \small
  \begin{tabular}{c | c c c }
    \toprule
    Dataset Split & Train Seen & Val Seen & Val Unseen \\
    \midrule
    \textrm{Task Domain Number} & $55$ & $53$ & $8$ \\
    \bottomrule
  \end{tabular}
\end{table}

For comparative experiments, we implemented the latest unpublished related work with the corresponding setup and a prior study incorporating continual learning. We also compared our approach with other VLN agents to validate performance on the test unseen split and confirm that continual learning enhances agent performance.

\subsection{Evaluation Protocol for VLNCL}


To assess the forgetting resistance and knowledge transfer abilities of the VLNCL agent, we introduce two metrics: \textit{Seen Transfer (ST)} and \textit{Unseen Transfer (UT)}, analogous to Backward and Forward Transfer in continual learning. \textit{ST} measures the average performance difference on the $i$-th task domain after training on $T$ domains compared to training on the first $i$ domains ($i < T$). \textit{UT} evaluates the average performance difference on unseen tasks after training on $T$ domains compared to the base agent. As training progresses, unseen tasks diminish as the agent encounters them. 
Note that the unseen task domain set is a subset of the validation task set, divided into several scenario-related data streams that feed the agent sequentially. Therefore, the unseen task domain set is dynamic.

Formally, the validation task set is separated into $d$ task domains $\mathcal{S}=\{s_1, s_2, ..., s_d\}$. The initial unseen task domain set is $\mathcal{S}_{unseen}^0=\{s_1, s_2, ..., s_d\}$. After the agent encounters task domain $s_1$, the unseen task domain set is updated with $\mathcal{S}_{unseen}^1=\{s_2, s_3, ..., s_d\}$. Hence, the Seen Transfer is defined as
\begin{equation}
    ST = \frac{1}{T-1} \sum_{i=1}^{T-1} SR_{T}(s_i) - SR_{i}(s_i) \quad s_i \in \mathcal{S}_{seen}^T
\end{equation}
The Unseen Transfer is defined as
\begin{equation}
    \resizebox{0.85\hsize}{!}{$
    \begin{aligned}
		UT = \frac{1}{T-1} \sum_{i=2}^{T} SR_{i-1}(s_i) - SR_{0}(s_i) \quad s_i \in \mathcal{S}_{unseen}^T
	\end{aligned}
  $}
\end{equation}
where the $SR_j(s_i)$ is the average success rate in the task domain $s_i$ for agent after trained on the $j$ task domain. The $SR_0(s_i)$ represents the success rate of the base agent.

\subsection{Implementation Details}



The agent's hyperparameters and the structured transformer architecture align with those utilized in previous studies. The candidate target grid size, $d$, is set to 5, resulting in a $5 \times 5$ grid with a spacing of 6 meters between adjacent positions. The agent follows the ground-truth path when applying teacher force under specific loss configurations. Conversely, actions are sampled from the predicted probability distribution during student forcing, leveraging distinct loss configurations. 
We train the base model using the Adam optimizer on two NVIDIA V100 GPUs for 100,000 iterations, employing a batch size of 20 and a learning rate of $1\times10^{-5}$ over 72 GPU hours.
Continual learning is subsequently conducted on two NVIDIA V100 GPUs for 1k iterations per task domain with learnable learning rates.

\subsection{Comparative Experiment Results}

To evaluate the task performance of our agent, we compared its results on the test unseen split against single-run performances of other VLN agents on the R2R dataset. The methods included in this comparison are Seq2Seq \cite{anderson2018vision}, SSM \cite{wang2021structured}, EnvDrop \cite{tan2019learning}, AuxRN \cite{zhu2020vision}, CCC \cite{Wang2022ccc}, PREVALENT \cite{hao2020towards}, AirBERT \cite{guhur2021airbert}, VLN$\circlearrowright$BERT \cite{hong2021bert} (init. OSCAR). 
The baseline used the base agent's performance before applying continual learning. We further compared the performance before and after continuous learning to demonstrate the potential of CL approaches.


\begin{table}[ht]
  \caption{
   We compare the single-run performance on the test unseen split of the R2R dataset. The last row presents the performance before and after applying continuous learning.
  }
  \centering
  \tabcolsep=0.21cm
  \small
  \begin{tabular}{c | c c c c c}
    \toprule
    Method & TL & NE $\downarrow$ & OSR $\uparrow$ & SR $\uparrow$ & SPL $\uparrow$\\
    \midrule
    \textrm{Human} & $11.85$ & $1.61$ & $90$ & $86$ & $76$	\\
    \midrule
    \textrm{Base(Before CL)} & 19.41 & 5.87 & 57 & 43 & 37 \\
    \midrule
    \textrm{Seq2Seq}  & 8.13 & 7.85 & 27 & 20 & 18 \\
    \textrm{SSM} & 20.40 & 4.57 & 50 & 61 & 46\\
    \textrm{EnvDrop}  & 11.66 & 5.23 & 59 & 51 & 47\\
    \textrm{AuxRN} & - & 5.15 & - & 55 & 51\\
    \textrm{CCC} & - & 5.30 & - & 51 & 48\\
    \textrm{PREVALENT} & 10.51 & 5.30 & 61 & 54 & 57\\
    \textrm{AirBERT}   & 10.51 & 5.30 & - & 54 & 51\\
    \textrm{VLN$\circlearrowright$BERT}   & 12.34 & 4.59 & - & 57 & 53\\
    \rowcolor{gray!40} \textrm{Dual-SR(Ours)}& 17.90 & 4.54 & 65 & 59 & 51\\
    \midrule
    \textrm{Performance Rise}   & - & -1.33 & 8 & 16 & 14\\
    \bottomrule
    \end{tabular}
\end{table}


The comparative results demonstrate that our method significantly enhances the task performance of the agent. The success rate increased by 16\%, and the oracle success rate increased by 8\% compared to the base agent, highlighting the substantial potential of a continual learning approach in VLN agents. Moreover, our results have achieved state-of-the-art performance in continual learning ability. Unlike agents that rely on elaborate environmental understanding mechanisms or meticulous fine-tuning, agents with CL capabilities can consistently enhance their performance in new scenes. However, integrating a more advanced inference architecture with the CL approach can significantly improve the agent's performance.


\subsection{Resisting Forgetting and Transferring Evaluation}



With continual learning, we introduce the concepts of Seen Transfer and Unseen Transfer for the VLN agent to evaluate its resistance to forgetting and ability to transfer knowledge. To assess forgetting resistance, we apply the average Seen Transfer to the validation train split and the validation seen split. We utilize the average Unseen Transfer on the validation unseen split to evaluate the agent's capability to transfer knowledge. As a baseline, we use the results of fine-tuning each task domain.

The Seen Transfer measures the agent's performance on the seen task domain set, $\mathcal{S}_{seen}^T$, within the current split, determining whether the agent retains its prior knowledge. In contrast, Unseen Transfer evaluates performance on the unseen task domain set, $\mathcal{S}_{unseen}^T$, within the current split, assessing knowledge transfer from previous tasks to current ones.

\begin{table}[ht]
    \centering
    \tabcolsep=0.39cm
    \small
    \begin{tabular}{c|cc|c}
        \toprule
        \multirow{2}{*}{\textrm{Method}} & \multicolumn{2}{c|}{\textrm{ST (Avg)}$\uparrow$} & \multirow{1}{*}{\textrm{UT (Avg)}$\uparrow$} \\ 
         & Train Seen & Val Seen & Val Unseen \\ 
         \midrule
         \textrm{Fine-tune} & -15.10 & -10.59 & -8.03 \\ 
         \midrule
         \rowcolor{gray!40} \textrm{Dual-SR} & -8.49 & -6.00 & 2.18 \\ 
         \bottomrule
    \end{tabular}
    \caption{Experiment results of the vision language navigation with continual learning agent in seen and unseen transfer.} 
    \label{tab:comparison}
\end{table}

\begin{figure}[ht]
\centering
\includegraphics[width=0.48\textwidth]{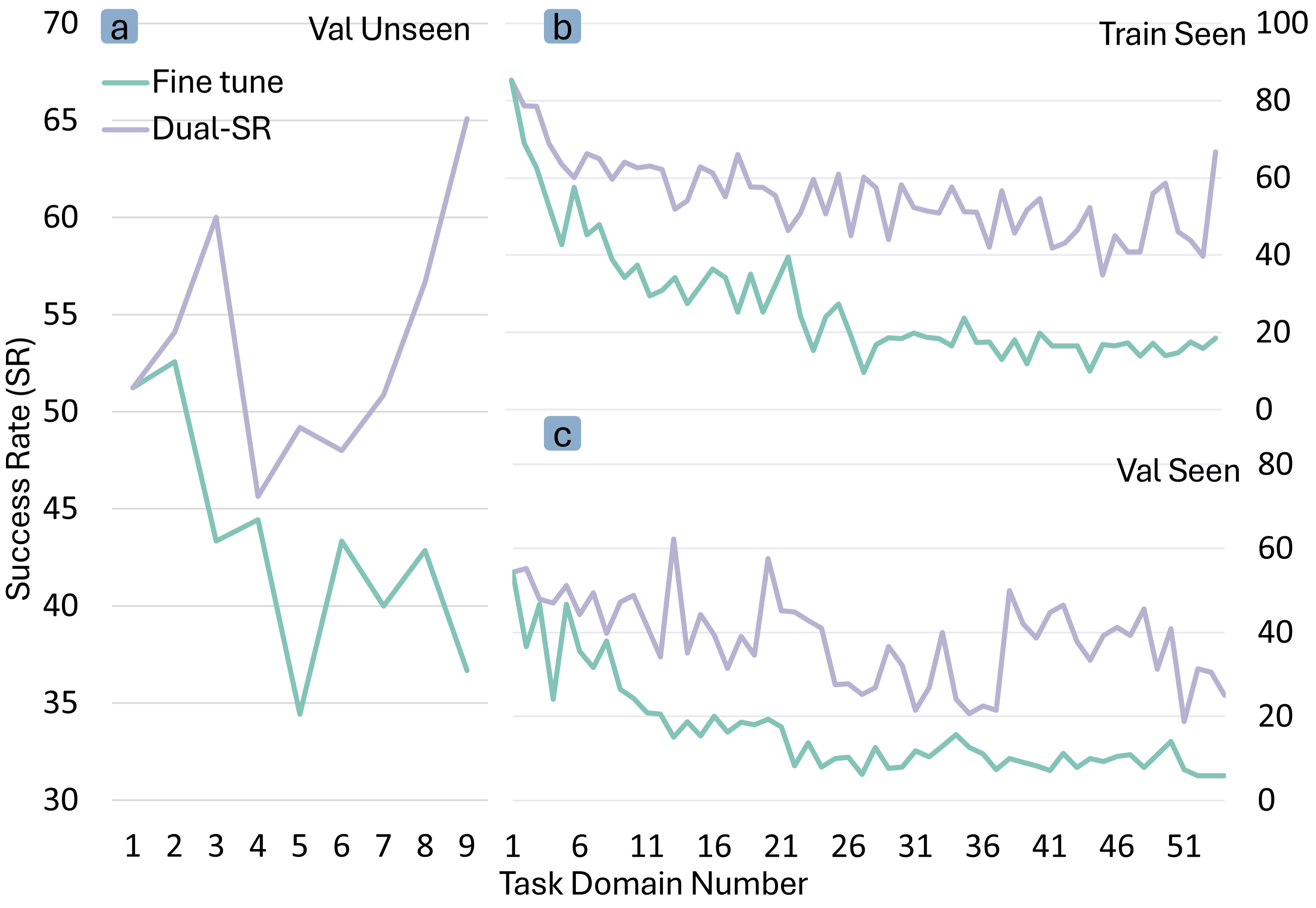}
\caption{The demonstration of success rate change while continuously learning new task domains. Part a is the performance in the Val Unseen split and evaluated on the unseen task domain set $\mathcal{S}_{unseen}^T$ to demonstrate the knowledge transfer capability. Parts b and c are performances in Train Seen and Val Seen splits on the seen task domain set $\mathcal{S}_{seen}^T$ to demonstrate the resistance ability to forget.}
\label{figure: change}
\end{figure}

The experimental results demonstrated that our method has advantages in forgetting resistance and knowledge transfer. To further elucidate these capabilities, we present the changes in success rates across different splits in Figure \ref{figure: change}. In the Val Unseen split, we evaluate the success rate in the unseen portion to assess knowledge transfer ability. Conversely, in the Val Seen and Train Seen splits, we consider the success rate in the seen portion to evaluate the agent's forgetting resistance.

Additionally, we can track the highest and lowest Success Rates (SR) and Oracle Success Rates (OSR) across all dataset splits in Table \ref{tab:max_and_min}. The evaluation of performance also follows the VLNCL setup. This result allows us to observe the changes in performance more clearly.

\begin{table}[ht]
    \centering
    \tabcolsep=0.12cm
    \small
    \begin{tabular}{c|c|cccccc}
        \toprule
        \multirow{2}{*}{\textrm{Method}} & \multirow{2}{*}{\textrm{Type}} & \multicolumn{2}{c}{\textrm{Train Seen}} & \multicolumn{2}{c}{\textrm{Val Seen}} & \multicolumn{2}{c}{\textrm{Val Unseen}} \\
        & & SR & OSR & SR & OSR & SR & OSR \\
        \midrule
        \multirow{3}{*}{\textrm{Fine-tune}} & Init & 85.27 & 89.27 & 54.31 & 61.37 & 51.24 & 62.69 \\
        & Max & 62.20 & 67.33 & 46.67 & 51.85 & 52.56 & 62.20 \\
        & Min & 9.64 & 16.22 & 5.83 & 13.51 & 34.43 & 38.82 \\
        \midrule
        \multirow{3}{*}{\textrm{Dual-SR}} & Init & 85.27 & 89.27 & 54.31 & 61.37 & 51.24 & 62.69 \\
        & Max & 78.67 & 86.60 & 69.00 & 71.74 & 65.08 & 73.59 \\
        & Min & 34.80 & 41.93 & 18.75 & 25.00 & 45.65 & 57.38 \\
        \bottomrule
    \end{tabular}
    \caption{Tracking the minimum and maximum Success Rates (SR) and Oracle Success Rates (OSR). 'Init' represents the agent's performance before continual learning.} 
    \label{tab:max_and_min}
\end{table}

Analyzing performance changes shows that our method exhibits strong resistance to forgetting and robust knowledge transfer capabilities. Our method consistently improves performance on unseen tasks in the Val Unseen split, highlighting the significant generalization capability introduced by continual learning in VLN agents. In the Val Seen and Train Seen splits, our method also shows substantial performance retention on seen tasks, with a success rate drop of only 20\% after processing a sequence of 50 different scenario tasks. These results indicate that the agent effectively mitigates forgetting across long-term task domains. Consequently, we propose a benchmark for the Vision-Language Navigation with Continual Learning paradigm. 
By implementing a continual learning approach in VLN agents, we enhance their ability to generalize effectively to unseen environments, paving the way for broader real-world applications of VLN agents.


\section{Conclusion}


This paper presents the Vision-Language Navigation with Continual Learning (VLNCL) paradigm, where the agent learns from unseen tasks while retaining knowledge from prior scenarios, closely reflecting real-world application demands. To achieve this, we introduce the Dual-loop Scenario Replay (Dual-SR) algorithm, which improves the agent's generalization and task performance. We also establish a benchmark for VLNCL using the R2R dataset. Experiments demonstrate that our method surpasses existing continual learning approaches under comparable conditions, advancing VLN agent performance and setting the stage for further research into real-world application-ready agents. Our future work will concentrate on developing more sophisticated inference mechanisms and advancing continual learning strategies to improve generalization.

\bibliography{aaai25}

\end{document}